\pdfoutput=1
\documentclass[11pt]{article}

\usepackage[preprint]{acl}
\usepackage{acl}

\usepackage{times}
\usepackage{booktabs}
\usepackage{latexsym}
\usepackage{ulem}
\usepackage{amsmath}
\usepackage[inline]{enumitem}
\usepackage{subcaption}

\usepackage{amsfonts}

\usepackage[T1]{fontenc}
\usepackage[utf8]{inputenc}

\usepackage{microtype}

\usepackage{inconsolata}

\usepackage{graphicx}
\usepackage{xcolor}

\begin{document}

\title{Enhancing Knowledge Graph Completion with GNN Distillation and Probabilistic Interaction Modeling}

% \title{An abstract feature method for knowledge graph completion using probabilistic feature interaction networks.}
% Author information can be set in various styles:
% For several authors from the same institution:
% \author{Author 1 \and ... \and Author n \\
%         Address line \\ ... \\ Address line}
% if the names do not fit well on one line use

% \author{First Author \\
%   Affiliation / Address line 1 \\
%   Affiliation / Address line 2 \\
%   Affiliation / Address line 3 \\
%   {\tt email@domain} \\\And
%   Second Author \\
%   Affiliation / Address line 1 \\
%   Affiliation / Address line 2 \\
%   Affiliation / Address line 3 \\
%   {\tt email@domain} \\}
  
\author{ Lingzhi Wang\textsuperscript{1, 2}, 
{\bf \textsuperscript{\textdagger}Pengcheng Huang\textsuperscript{2}},
{\bf Haotian Li\textsuperscript{1, 2}},
{\bf Yuliang Wei\textsuperscript{1, 2}}, \\ 
{\bf Guodong Xin\textsuperscript{1, 2}},
{\bf Rui Zhang\textsuperscript{1, 2}}, 
{\bf Donglin Zhang\textsuperscript{1, 2}}, 
{\bf Zhenzhou Ji\textsuperscript{2}}, 
{\bf \textsuperscript{*} Wei Wang\textsuperscript{1, 2}},\\
\textsuperscript{*} Corresponding Author,
      \textsuperscript{\textdagger} Co-first Author\\
      \textsuperscript{1} Shandong Key Laboratory of Industrial Network Security  \\
      \textsuperscript{2} School of Computer Science and Technology, Harbin Institute of Technology, Weihai 264209,
\\
 \small{
   \textbf{Correspondence:} \texttt{\{23s130410, 23s030138, 24b903122, 24s030150\}@stu.hit.edu.cn, 
 }}
\\
 \small{
    \texttt{\{gdxin, wei.yl, jizhenzhou,wwhit\}@hit.edu.cn},
   \texttt{lihaotian1231@gmail.com},
 }
% \\
% \small{
%   \texttt{\{23s130410, 23s030138, 24b903122, 24s030150\}@stu.hit.edu.cn}, \\
%   \texttt{\{gdxin, wei.yl, jizhenzhou,wwhit\}@hit.edu.cn}, \\
%   \texttt{lihaotian1231@gmail.com}
}

\maketitle
\begin{abstract}
  % Knowledge graphs (KGs) are crucial structures that organize large-scale, interconnected
  % data in a variety of domains, from semantic search and natural language understanding to 
  % recommendation systems. Despite their utility, most KGs remain incomplete, limiting their 
  % full potential in downstream applications. Knowledge graph completion (KGC) seeks to address 
  % this by predicting missing links, relying heavily on accurate modeling of both local and 
  % global relational patterns. However, existing methods face two major challenges: deep graph 
  % neural networks (GNNs) often suffer from over-smoothing as layer depth increases, and 
  % pre-trained language models, though adept at extracting semantic information from entity 
  % descriptions, struggle to capture higher-dimensional, abstract relational features crucial 
  % for effective link prediction.
% Knowledge graphs (KGs) serve as foundational structures for organizing interconnected data 
% across diverse domains. However, most KGs remain incomplete, significantly 
% limiting their effectiveness in downstream applications. Knowledge graph completion (KGC) 
% addresses this by inferring missing links, 
Knowledge graphs (KGs) serve as fundamental structures for organizing interconnected data 
across diverse domains. However, most KGs remain incomplete, limiting their effectiveness 
in downstream applications. Knowledge graph completion (KGC) aims to address this issue by 
inferring missing links,
but existing methods face critical challenges: deep graph neural networks (GNNs) suffer 
from over-smoothing, while embedding-based models fail to capture abstract relational features.
This study aims to overcome these limitations by proposing a unified framework that integrates 
GNN distillation and abstract probabilistic interaction modeling (APIM). GNN distillation 
approach introduces an iterative message-feature filtering process to mitigate over-smoothing, 
preserving the discriminative power of node representations. APIM module complements this 
by learning structured, abstract interaction patterns through probabilistic signatures and 
transition matrices, allowing for a richer, more flexible representation of entity and relation 
interactions. We apply these methods to GNN-based models and the APIM to embedding-based KGC models, 
conducting extensive evaluations on the widely used WN18RR and FB15K-237 datasets. Our results demonstrate 
significant performance gains over baseline models, showcasing the effectiveness of the 
proposed techniques. The findings highlight the importance of both controlling information 
propagation and leveraging structured probabilistic modeling, offering new avenues for 
advancing knowledge graph completion. And our codes are available at \url{https://anonymous.4open.science/r/APIM_and_GNN-Distillation-461C}

% \footnote{Our codes are available at \url{https://github.com/Wlz-Hit/APIM_and_GNN-Distillation}.}
\end{abstract}

\section{Introduction}

Knowledge graphs (KGs)\cite{pan2017exploiting} are a type of semantic representation 
of the knowledge in the form of triples, where each triple consists of a subject, 
a predicate, and an object. Large-scale KGs, such as Wikidata\cite{vrandevcic2014wikidata}, 
DBpedia\cite{lehmann2015dbpedia}, and Freebase\cite{bollacker2008freebase}, have 
been generated and successfully applied in various applications such as question 
answering\cite{cui2023incorporating,wang2023novel}, information retrieval\cite{fei2021enriching,dutkiewicz2024knowledge}, 
and natural language processing\cite{hu2024geoentity}. However, most of the KGs 
are incomplete\cite{bollacker2008freebase}, and they need to be completed to enable
various applications. Therefore, knowledge graph completion (KGC) is the task of 
predicting missing facts or triples in a knowledge graph (KG) based on the existing 
facts or triples\cite{shen2022comprehensive}. The most important step for KGC better
performance is to capture the suitable characteristics of the entities and relations in 
the KGs. 

Recently, there have been several works on KGC, such as the Graph Neural Networks (GNNs)
models based the intrinsic graph-structure of KGs\cite{li2023message} and the embedding 
models, particularly large-scale language model (LLM) based methods generated the KGs 
semantic representation\cite{wang2022simkgc}. 
% 补充引用文献

The GNNs-based modules iteratively aggregating information from neighboring nodes to learn low-dimensional representations
of entities and relations. Unlike traditional embeddings approaches (e.g., TransE\cite{bordes2013translating}
ComplEx\cite{trouillon2016complex}, Tucker\cite{balavzevic2019tucker}) for KGs, the GNNs-based
methods take advantage of the topology of the knowledge graph, thereby enhancing the representations
with contextual information\cite{hamilton2017inductive}. In the context of knowledge graph completion,
a widely recognized challenge is to capture and integrate relational information while preserving the 
the discriminative features of entities. For instance, the Relational Graph Convolutional Network 
(RGCNs)\cite{schlichtkrull2018modeling}. By modeling each relation type separately, RGCNs can better
capture the semantic nuances across different types of edges. More broadly, GNN variants such as 
Graph SAGE\cite{hamilton2017inductive} and Graph Attention Networks (GATs)\cite{velivckovic2017graph, nathani2019learning} 
have also been adapted for KGs, offering inductive and attention-based mechanisms to aggregate neighborhood entities' features. 

\textbf{Problem 1.} A critical limitation of deep GNNs in KGs applications is over-smoothing, where 
repeated message-passing operations cause node representations to become indistinguishable, degrading 
model performance\cite{li2018deeper, oono2019graph}.
% Over-smoothing arises because GNNs iteratively average neighborhood 
% features, causing node representations to converge to a low-dimensional subspace. Researchers\cite{oono2019graph}
% proved that the rank of node representations decays exponentially with GNN depth, leading to information loss.

The LLM-based methods adapt the pre-trained language model (e.g., BERT\cite{devlin2019bert}, 
ChatGpt\cite{nazir2023comprehensive}) to leverage the semantic information of extra entities description.
In these method, each triple $(h, r,t)$ is converted into a textual sequence, which often 
by concatenating the textual description of the head entity $h$, the relation $r$, and the
tail entity $t$. The combined sequence is then fed into the pre-trained language model to 
produce a contextual embedding, which can be used to predicate the missing triples.

\textbf{Problem 2.} Although pre-trained language models can effectively capture the semantic 
nuances of auxiliary entity descriptions, they fall short in modeling the higher-dimensional, 
abstract features necessary for knowledge graph completion. In essence, these models do not 
adequately represent the sophisticated, latent relational patterns—beyond mere local and 
global structures—that are critical for capturing the intricate interplay between entities 
and relations in knowledge graphs.

\textbf{Present work.} In this paper, we propose a GNN distillation method and an 
abstract probabilistic interaction modeling (APIM) method to address the aforementioned 
challenges. In contrast to standard distillation approaches, our GNN distillation 
method incorporates an iterative message-feature filtering process at each GNN layer. 
By filtering and propagating only the most informative features, this technique 
effectively mitigates over-smoothing, thereby enhancing the embeddings’ representational 
capacity and preserving higher-order relational cues within the knowledge graph. In parallel, 
our abstract features capture method models the interaction probability of abstract 
features across different entities and computes the expected value of these feature-
mediated interactions, thereby capturing both local and global structural information. 
Specifically, we apply both methods to traditional GNN-based KGC models and employ the 
abstract features capture method in traditional embedding-based KGC models. Empirical 
evaluations demonstrate that our proposed approaches outperform their respective 
baselines on knowledge graph completion tasks.

% \textbf{Contribution.} This work makes three key contributions:
% \begin{itemize}
%   \item We propose a GNN distillation method that incorporates iterative message-feature 
%   filtering, effectively mitigating over-smoothing and improving the representational 
%   capacity of embeddings.
%   \item We introduce an Abstract Probabilistic Interaction Modeling (APIM) method that 
%   models feature interactions across entities, capturing both fine-grained and abstract 
%   structural information.
%   \item We conduct empirical evaluations demonstrating that our methods surpass 
%   existing baselines in knowledge graph completion tasks.
% \end{itemize}
\section{Preliminaries}
% Before diving into the details of our proposed method, we first introduce some KGC
% preliminaries.

\subsection{Knowledge graph completion (KGC)}
Knowledge Graph Completion (KGC) constitutes a pivotal task in knowledge representation 
learning, dedicated to predicting missing connections between entities within incompletely 
observed knowledge graphs (KGs). Formally, given a KG defined as 
$\mathcal{G}=\{(h, r, t)\} \in \mathcal{E} \times \mathcal{R} \times \mathcal{E}$, where 
$\mathcal{E}$ and $\mathcal{R}$ denote the entity set and relation type set respectively, 
the KGC task involves inferring plausible triples for masked positions in either head or tail
entity slots\cite{shen2022comprehensive}, expressed as $(h, r, \cdot)$ or $(\cdot, r, t)$.
The standard evaluation protocol requires the model to rank all candidate entities $\mathcal{E}$
based on a learned scoring function $f : \mathcal{E} \times \mathcal{R} \times \mathcal{E} \to \mathbb{R}$,
where $\mathbb{R}$ is a real-valued score.
For a given query $(h, r, \cdot)$ or $(\cdot, r, t)$, the model computes the plausibility
$f(h, r, t')$ or $f(h', r, t)$ for each candidate entity $t' \in \mathcal{E}$ or $h' \in \mathcal{E}$,
respectively. The candidate with maximal score is selected as the optimal completion, following 
the closed-world assumption\cite{oh2022open}.

% Knowledge Graph Completion (KGC) is a fundamental task that aims to infer missing 
% links among head or tail entities in a Knowledge graph (KG). Specifically, given a 
% knowledge graph $\mathcal{G}=\{(h, r, t) \}$, where $h$ (head) and $t$ (tail) are 
% entities and $r$ is a relation type, KGC masks indicates the missing triples denoted
% as $(h, r, \cdot)$ or $(\cdot, r, t)$. Then use all entities in $\mathcal{G}$ as 
% candidates and try to select the most suitable ones to fill the missing triples.
% Typically, for each candidate entity $t'$ or $h'$, we can compute a score function
% $f(h, r, t')$ or $f(h', r, t)$ to measure the likelihood of the triple $(h, r, t')$
% or $(h', r, t)$ being true. The candidate entity with the highest score is then 
% selected as the most suitable one to fill the missing triple. The function $f(\cdot)$
% can be modeled in various ways, such as logistic regression, cosine similarity,
% neural networks and so on. 

\subsection{GNN-based KGC methods}
The GNN-based KGC methods have emerged as a dominant paradigm in KGC, primarily 
categorized into two types: Message Passing Neural Networks (MPNNs) and Path-based
methods. In this work, we focus on the MPNNs-based methods, which operate on the 
principle of iterative neighborhood aggregation, where entity representations are 
updated by propagating an transforming information from connected neighbors. 

\begin{figure}[!h]
  \includegraphics[width=\columnwidth]{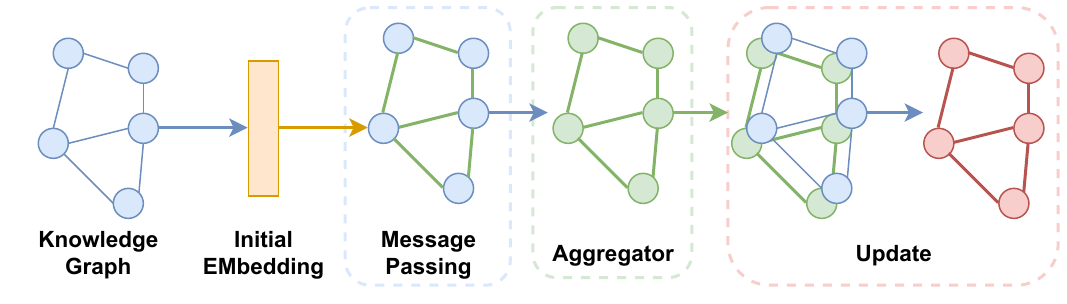}
  \caption{A general structure of the MPNNs-based KGC methods.}\label{fig:MPNNS_Structure}
\end{figure}

We demonstrate the general structure of the MPNNs-based KGC methods in Figure~\ref{fig:MPNNS_Structure}. 
This framework is formally defined by three components:
\begin{equation}
\begin{split}
  &Message: m_{ij}^{(l)} = \phi\left(h_i^{(l-1)},h_j^{(l-1)},r_{ij}\right) \\
  &Aggregation: M_i^{(l)}=\bigoplus_{j\in\mathcal{N}(i)}m_{ij}^{(l)} \\
  &Update: h_i^{(l)}=\psi\left(h_i^{(l-1)},M_i^{(l)}\right)
\end{split}
\label{equ:MPNN}
\end{equation}
where $\phi$, $\bigoplus$, and $\psi$ denote message functions, permutation-invariant 
aggregator, and update functions, respectively.

\subsection{Emebdding-based KGC methods}

% Embedding-based approaches to KGC learn low-dimensional vector representations for 
% entities and relations in a shared latent space. Most embedding-based KGC methods 
% follow a similar training pipline. First, each entity and relaiton is intialized 
% with a learnable embedding (either randomly or by pre-trained step). Then, a margin-based 
% or cross-entropy loss is employed to separate positive triples from negative samples.
% Negative sampling typically involves replacing either head or tail entities with random
% entities not observed in the original triples. Once trained, inference is performed by 
% ranking all candidate entities according to the model's scoring function and selecting
% the highest-scoring entity as the predicted head or tail. Typically, they can be grouped 
% into three main categories: translational distance models, bilinear Models, neural 
% network-based models. Neural network-based KGC models offer increased flexibility and
% expressiveness, compared to purely translational or bilinear models. Although potenially
% more computationally demanding, their capacity to represent intricate relational patterns
% often leads to improved performance on complex or large-scale KGs. Besides, the pre-training 
% language model for KGC is a model that can be trained on a 
% large amount of text data and can generate high-quality text that can be used as 
% input to the KGC method and has more powerful ability to KGC.Therefore, in this 
% paper, we focus on the neural network-based models, which are the most recent 
% innovations in KGC.

Embedding-based approaches to KGC learn low-dimensional representations for both 
entities and relations within a shared latent space. Typically, these methods adhere 
to a common training pipeline: each entity and relation is first initialized with 
a learnable embedding, either randomly or via pre-training; thereafter, a margin-based 
.or cross-entropy loss is employed to differentiate positive triples from negative 
samples generated through random replacement of head or tail entities. Inference 
is performed by ranking all candidate entities according to the model's scoring 
function and selecting the highest-scoring candidate as the prediction. These 
approaches can be broadly classified into three categories: translational 
distance models, bilinear models, and neural network–based models. Neural 
network–based models, despite being potentially more computationally intensive, 
offer greater flexibility and expressiveness, enabling them to capture intricate 
relational patterns and achieve superior performance on complex or large-scale 
knowledge graphs. Furthermore, pre-trained language models have been leveraged 
in KGC to generate high-quality textual representations that serve as valuable 
inputs, further enhancing model capabilities. Consequently, in this paper, 
we focus on neural network–based models, which represent the state-of-the-art 
in knowledge graph completion.

\section{Methodologys}
% We propose a GNN distillation method and an abstract features capture method to 
% address over-smoothing and abstract feature incompleteness, respectively.

\subsection{GNN distillation method}
Traditional GNNs often rely on iterative message passing, where each node (entity) 
aggregates information from its neighbors. As we increase the number of layers, 
representations of different entities may converge to similar values\textemdash 
an effect known as over-smoothing. The GNN distillation method is designed to tackle 
this problem by introducing a feature distillation step that filters and refines 
message signals in each GNN layer, ensuring entities maintain more discriminative 
embeddings. 

We start with a standard GNN architecture (e.g., GCN, GAT, RGCN). During each 
forward pass at layer $l$, every node $i$ receives a message from its neighbors
$\mathcal{N}(i)$ and aggregates them as the Equation~\ref{equ:MPNN}. After receiving 
the initial aggregated messages, a feature distillation operator is applied to 
filter out less informative and dampens or eliminates features that contribute to 
over-smoothing. Formally, let $m_{i}^{(l)}$ denote the message for node $i$ at 
layer $l$. The operator outputs a distilled message $\widetilde{m}_i^l$ based on 
an importance score:
\begin{equation}
  \widetilde{m}_i^{(l)} = Distill(m_{i}^{(l)}; \alpha_i^{(l)})
\label{equ:Distillation}
\end{equation}
where $\alpha_i^{(l)}$ are parameters that control the distillation ratio.
The distillation process repeats multiple rounds within the same layer. Each round 
reevaluates the messages, removing or down-weighting redundant features and retaining
only those deemed most important. Mathematically, at the $k$-th iteration of distillation (where $k \in \{1, \cdots , \mathbb{K}\}$)
with layer $l$, the refined message is computed as:
\begin{equation}
  \widetilde{m}_{i}^{(l)} = Distill(\widetilde{m}_i^{(l, k-1)}; \alpha_i^{(l, k)})
\label{equ:DistillationRound}
\end{equation}
with $\widetilde{m}_i^{(l, 0)}=\widetilde{m}_i^{(l)}$ serving as the intial input for
round 1. To ensure we gradially reduce the amount of message information and avoid 
abruptly dropping too much signal, we adopt a linear exponential decay schedule that 
determines how many features are retained at each round:
\begin{equation}
\alpha^{(k)}=\begin{cases}
                \alpha_{start} - (k - 1)\Delta, & \text{(linear \ decay)} \\
                \alpha_{start} \cdot \gamma^{(k-1)}, & \text{(exponential \ decay)}
              \end{cases}
\label{equ:distillation_schedule}
\end{equation}
where $\alpha^{k}$ indicates the proportion of features retained or the threshold
for importance scoring at round $k$, and $\Delta$ and $\gamma$ are hyperparameters.
This progressive filtering keeps the most critical features while discarding increasingly
unhelpful or redundant signals. The node representation then incorporates the distilled 
message $\tilde{m}_i^{(l, k)}$ (i.e., after the final round of distillation), update node
representations via:
\begin{equation}
  h_i^{(l)}=\gamma\left(h_i^{(l-1)},\tilde{m}_i^{(l, K)}\right)
  \label{equ:Update}
\end{equation}

\begin{figure}[t]
  \includegraphics[width=\columnwidth]{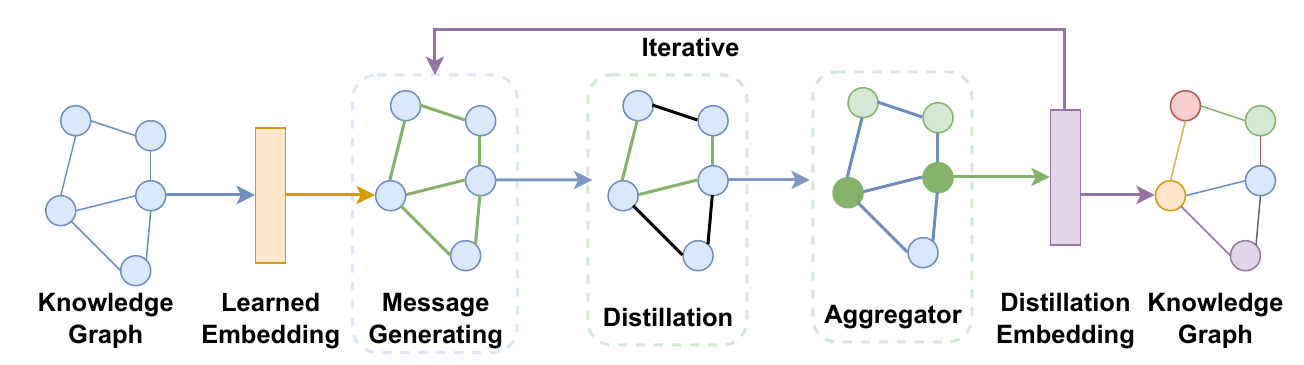}
  \caption{The visualisation of the GNN Distillation method.}
  \label{fig:GNN_Distillation}
\end{figure}

As shown in Figure~\ref{fig:GNN_Distillation}, by filtering out repetitive and 
uninformative message components, GNN distillation preserves feature diversity 
across different entities. This helps nodes maintain more discriminative features, 
rather than converging to nearly identical vectors. The iterative message filtering 
ensures that nodes still benefit from aggregated information (global context) but 
do not over-absorb neighbor features to the point of losing local uniqueness. 

The distillation operator can be 
integrated into various GNN frameworks (GCN, GAT, RGCN) by adjusting the 
filtering mechanism or the decay schedule. It minimally impacts the model’s 
overall complexity while offering robust control over message propagation.

\subsection{{The abstract probabilistic interaction modeling (APIM) method}}
APIM introduces a probabilistic interaction paradigm that explicitly models 
entity-relation interactions through learnable abstract patterns. For each 
entity $e \in \mathcal{E}$, we learn a probabilistic interaction signature:
\begin{equation}  
  \mathbf{a}_e=\sigma\left(\mathbf{W}_a\mathbf{h}_e\right)\in[0, 1]^K
  \label{equ:Abstract_signature}
\end{equation}
where $\sigma$ is the sigmoid function, $h_e \in \mathbb{R}^d$ is the entity 
embedding, $\mathbf{W}_a$ is a trainable projection matrix, and $K$ denotes 
the number of latent interaction modes. This step encodes how entities 
probabilistically engage in different interaction modes.

To focus on the most salient interaction patterns, we retain only the top-k 
modes for each entity:
\begin{equation}
\mathbf{\widetilde{a}}_e = \text{TopK}(\mathbf{a}_e, k) \odot \mathbf{a}_e
\label{equ:topk_mask }
\end{equation}
where $\text{TopK}(\cdot, k)$ generates a binary mask preserving the indices of the $k$
largest values in $\alpha_e$, and $\odot$ denotes element-wise multiplication. 
This sparsification reduces noise from low-probability modes while maintaining 
gradient flow through the masked values.

In parallel, for each relation 
$r \in \mathcal{R}$ maintains a matrix:
\begin{equation}
  \quad\mathbf{P}_r[i,j]=P(\mathrm{mode}_i\to\mathrm{mode}_j\mid r)
\end{equation}
where $\mathbf{P}_r\in\mathbb{R}^{K\times K}$ represent transition probabilities between 
interaction modes conditioned on relation $r$. Matrices are normalized via tanh:
\begin{equation}
  \mathbf{P}_r=\mathrm{tanh}(\mathbf{\Theta}_r),\quad\mathbf{\Theta}_r\in\mathbb{R}^{K\times K}
\end{equation}
where $\mathbf{\Theta}_r$ is the trainable projection matrix of relation $r$. 
The values of $\mathbf{P}_r$ represent the positive and negative probabilistic 
interactions between entity modes, respectively. 

The triple score is then computed as the expectation of these interactions, combining 
the abstract interaction vectors with the transition matrix:

\begin{equation}
  f(h,r,t)=\mathbf{\widetilde{a}}_h^T\mathbf{P}_r\mathbf{\widetilde{a}}_t
  \label{equ:Abstractive_score}
\end{equation}
This formulation models entity interactions as weighted combinations of latent 
relational patterns, providing a flexible framework for capturing complex relational dynamics.

During training, the APIM module is trained in an end-to-end manner with the loss 
function:
\begin{equation}
  \begin{split}
  \mathcal{L}_{\mathrm{APIM}} = & -\sum_{(h,r,t)\in\mathcal{T}} \left[ y_{hrt} \log \sigma(f_{\mathrm{APIM}}(h,r,t)) \right. \\
  & \left. + (1-y_{hrt}) \log(1-\sigma(f_{\mathrm{APIM}}(h,r,t))) \right] \\
  & + \lambda \cdot |\mathbf{P}_r|_F^2,
  \end{split}
  \label{equ:APIM_loss}
\end{equation}
where $y_{hrt}$ denotes the ground-truth label for the triple, and $\sigma(\cdot)$ represents 
the sigmoid function. The term $\lambda \cdot |\mathbf{P}_r|_F^2$ prevents overfitting by 
constraining the magnitude of probability transition matrices, maintains numerical stability 
and approximates the effect of implicit distribution alignment through gradient-driven learning.

Generally, the TopK selection aligns with the \textit{information bottleneck principle} 
\cite{cao2023justicesinformationbottlenecktheory}, filtering irrelevant interaction modes 
to prevent overfitting. Although the learned sparse signatures $\mathbf{\widetilde{a}}_e$ 
do not provide immediate interpretability of the underlying interaction patterns, they 
effectively model latent modes that capture complex relational structures. This abstraction 
enables the framework to generalize across diverse knowledge graph domains, providing a 
systematic approach to handle high-dimensional interactions. 
% By focusing on latent rather 
% than explicitly defined modes, the APIM approach introduces a novel layer of abstraction 
% that sets it apart from traditional methods, offering a more flexible and robust modeling 
% capability.

% Generally, the Topk selection aligns with the \textit{information bottleneck principle}
% \cite{cao2023justicesinformationbottlenecktheory}, filtering irrelevant interaction modes 
% to prevent overfitting. Sparse signatures $\mathbf{\widetilde{a}}_e$ allow explicit interpretation 
% of dominant entity behaviors.

%(see case study in Section~\ref{appendix:case_study}).

% \textcolor{red}{For $K\geq|\mathcal{P}|$ where $\mathcal{P}$ is the set of ground-truth interaction patterns,
% APIM can recover all valid entity-relation interactions with probability $1-\epsilon$ under 
% proper initialization (derived via universal approximation properties of 
% softmax mixtures). Besides, the probabilistic transition matrix $\mathbf{P}_r$ induces a 
% relation-aware convex hull in the interaction space, enable adaptive pattern selection across
% relations. The hull dimensionality satisfies $\text{rank}(\mathbf{P_r}) \leq \text{min}(K, |\mathcal{P}|)$
% , where $\mathcal{P}$ is the pattern set for relation $r$}.

\subsection{The GNN-based models enhancement}

% Building upon the detailed descriptions of the GNN distillation and abstract probabilistic 
% interaction modeling (APIM) methods presented above, we now describe how these mechanisms 
% are integrated into existing GNN-based knowledge graph completion (KGC) frameworks—specifically 
% KB-GAT, RGCN, and ComPGCN—to enhance their performance. This section elucidates the 
% architectural modifications, training strategy, and loss computation enhancements that jointly 
% contribute to improved representation learning.

\begin{figure*}[!h]
  \includegraphics[width=\linewidth]{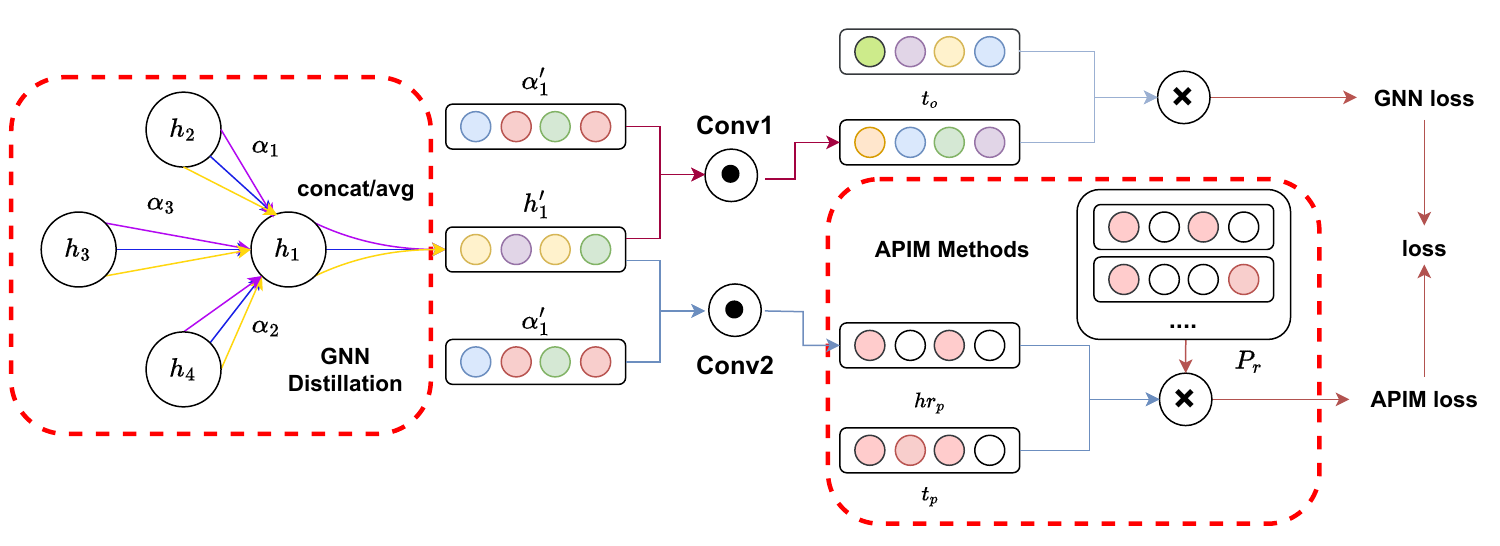}
  \caption{The enhanced KB-GAT framework integrates a GNN distillation module 
  that refines message representations by combining them with attention weights 
  $\alpha$ and an APIM module that models abstract interaction patterns. 
  In the diagram, $\otimes$ denotes matrix 
  multiplication, while $\odot$ represents the GCN aggregation operation.}
  \label{fig:GNN_based_model_framework}
\end{figure*}

In traditional models such as KB-GAT, RGCN, and ComPGCN, node embeddings are iteratively 
refined through neighborhood aggregation. Our GNN distillation method is incorporated into 
each GNN layer to counteract the over-smoothing problem. For instance, in KB-GAT the attention
mechianism is modified using a distillation operator into the message-passing process as shown
in Figure~\ref{fig:GNN_based_model_framework}. Concretely, before computing the attention 
weights as in \cite{velivckovic2017graph}, the neighbor messages $m_{ij}^{(l)}$ are subjected 
to an iterative filtering process that yields distilled messages $\tilde{m}_i^{(l, K)}$.
These distilled messages are then used in place of the original aggregated information, 
ensuring that only the most discriminative features are propagated. Similarly, in RGCN 
\citep{schlichtkrull2018modeling}, which aggregates information in a relation-specific 
manner, the distillation operator refines the aggregation process by dynamically 
attenuating redundant features. In ComPGCN, where compositional interactions are 
central to capturing higher-order relational patterns, the distillation process 
further enhances the model's ability to preserve interaction-specific signals 
without succumbing to the over-smoothing that often plagues deeper layers 
\citep{li2018deeper, oono2019graph}.

% \begin{figure*}[!h]
%   \includegraphics[width=\linewidth]{./images/GNN_based_model_framework.pdf}
%   \caption{The enhanced KB-GAT framework integrates a GNN distillation module 
%   that refines message representations by combining them with attention weights 
%   $\alpha$ and a APIM module that models abstract interaction patterns. 
%   In the diagram, $\otimes$ denotes matrix 
%   multiplication, while $\odot$ represents the GCN aggregation operation.}
%   \label{fig:GNN_based_model_framework}
% \end{figure*}

Concurrently, the APIM module is appended to the architecture following the final GNN 
layer. Each entity’s distilled embedding is transformed into an abstract probabilistic 
signature as Equation~\ref{equ:Abstract_signature}, where each relation $r$ is endowed
with a trainable transition matrix $\mathbf{P}_r$ that encodes inter-modal interactions.
The triple score is computed as Equation~\ref{equ:Abstractive_score}, thereby complementing
the local structural information captured by the GNN with global abstract interaction patterns.

The enhanced models are trained under a joint optimization framework that balances 
the conventional GNN-based loss with an auxiliary loss imposed by the APIM module. 
Specifically, the overall loss function is formulated as:
\begin{equation}
  \mathcal{L}=\mathcal{L}_\mathrm{GNN \ Distillation}+\lambda_\mathrm{APIM}\mathcal{L}_\mathrm{APIM}
\end{equation}
% where $\mathcal{L}_\mathrm{GNN \ Distillation}$ is the standard margin-based or
% cross-entropy loss used in enhancement GNN models (KB-GAT, RGCN, and ComPGCN) 
% for discriminating positive triples from negative samples, $\mathcal{L}_\mathrm{APIM}$
% is an auxiliary loss—implemented as cross entropy as Equations~\ref{equ:APIM_loss}—that enforces 
% consistency between the learned abstract interaction signatures and the expected 
% relational patterns and $\lambda_\mathrm{APIM}$ is a hyperparameter that modulates 
% the contribution of the APIM component to the overall loss.
where $\mathcal{L}_\mathrm{GNN \ Distillation}$ represents the standard margin-based 
or cross-entropy loss used in enhanced GNN models (KB-GAT, RGCN, and ComPGCN) to 
discriminate positive triples from negative samples. The auxiliary term $\mathcal{L}_\mathrm{APIM}$, 
defined as cross-entropy loss (Equation~\ref{equ:APIM_loss}), aligns the learned abstract 
interaction signatures with expected relational patterns. The hyperparameter $\lambda_\mathrm{APIM}$ 
controls the relative contribution of the APIM component to the overall loss.

By integrating the GNN distillation into the KB-GAT, RGCN, and ComPGCN frameworks, our approach 
effectively mitigating over-smoothing by preserving discriminative local features through 
iterative message filtering.

\subsection{The Embedding-based models enhancement}
In addition to the improvements in GNN-based KGC methods, we extend the Abstract 
Probabilistic Interaction Modeling (APIM) mechanism to enhance recent state-of-the-art 
embedding-based methods for knowledge graph completion. Embedding-based approaches, 
such as TransE \citep{bordes2013translating} and ComplEx \citep{trouillon2016complex}, 
typically learn fixed vector representations for entities and relations by optimizing 
a scoring function. Our enhancement integrates APIM into these models to capture 
abstract interaction patterns, thereby enriching the learned representations with 
additional structural and semantic cues. Within the embedding-based paradigm, each 
entity embedding $h_e$ is transformed into an abstract probabilistic signature via
Equation~\ref{equ:Abstract_signature}. Each relation $r$ is assigned a transition 
matrix $\mathbf{P}_r$, which encodes the latent interaction modes between entities.
The APIM score for a given triple $(h,r,t)$ is computed as Equation~\ref{equ:Abstractive_score}.
This score is designed to complement the original embedding-based scoring function by
explicitly modeling the probabilistic interactions underlying entity-relation pairs.

The overall training objective for the enhanced embedding-based model combines the 
original loss function $\mathcal{L}_{\mathrm{Emb}}$ (which is typically formulated 
as a margin-based or cross-entropy loss \citep{bordes2013translating, trouillon2016complex}) 
with an auxiliary loss component associated with the APIM module. In our framework, 
the APIM-derived score is directly incorporated into the loss computation as the 
Equation~\ref{equ:APIM_loss}. The overall loss is then defined as:
\begin{equation}
  \mathcal{L}=\mathcal{L}_\mathrm{Emb}+\lambda_{\text{APIM}}\mathcal{L}_\mathrm{APIM}
\end{equation}
with $\lambda_{\text{APIM}}$ serving as a hyperparameter that controls the contribution of the APIM module
relative to the original embedding-based loss. Through backpropagation, the gradients computed 
from this combined loss update both the original embedding parameters and the APIM-specific 
parameters (i.e., $\mathbf{W}_a$ and $\mathbf{P}_r$), thereby enabling the model to jointly 
optimize for traditional embedding fidelity and abstract interaction modeling.

By integrating the APIM mechanism, the embedding-based methods benefit from an enriched 
representation space that more effectively captures complex relational patterns. 
% The APIM component supplements the original score with a nuanced view of entity interactions, 
% contributing to improved discrimination between positive and negative triples.

Empirical results (detailed in Section~\ref{sec:Experiments}) consistently show 
that our enhanced model surpasses its baseline counterparts, thereby validating 
the effectiveness of integrating APIM into embedding-based methods and GNN 
distillation into GNN-based knowledge graph completion frameworks.

\section{Experiments}\label{sec:Experiments}
% We validate the effectiveness of our proposed GNN Distillation and Abstract 
% Probabilistic Interaction Modeling (APIM) methods on two standard knowledge 
% graph completion benchmarks: WN18RR and FB15K-237. All experiments are conducted 
% under a unified framework to ensure fair comparison.

\subsection{Experiments setup}
\textbf{Dataset}  WN18RR \cite{toutanova2015observed} and FB15K-237 \cite{dettmers2018convolutional} are 
widely adopted for evaluating relational reasoning capabilities. WN18RR contains 
hierarchical lexical relations with high semantic ambiguity, while FB15K-237 
features diverse real-world relations with significant long-tail distributions. 
Statistics are summarized in Appendix~\ref{appendix:dataset}.

\textbf{Evaluation Metrics} Following standard practice in KGC\cite{bordes2013translating},
our models are evaluated using the entity ranking task. For each test triple $(h, r, t)$, 
we perform tail entity prediction by ranking all candidate entities given $h$ and $r$ and 
similarly for head entity prediction. Subsequently, we evaluate the prediction performance 
using rank-based metrics—specifically, the Mean Reciprocal Rank (MRR) and Hits@N. In our 
experiments, we consider $N \in \{1, 3, 10\}$ for model evaluation. Detailed definitions 
of MRR and Hits@N are provided in Appendix~\ref{appendix:metrics}.

\textbf{Hyperparameters}
For GNN-based models, our framework employs a 4-layer GNN architecture with 100-dimensional initial embeddings 
expanded to 200-dimensional hidden representations. To balance model depth and assess the 
impact of GNN distillation on mitigating over-smoothing, we implement a linear decay 
distillation schedule (see Appendix~\ref{appendix:decay_selection} for experimental details) 
as defined by Equation~\ref{equ:distillation_schedule}, using $\alpha_{start}=1.0$, $k \in [2, 5]$, and $\Delta=0.2$.
For all models, the APIM module utilizes
100-dimensional probabilistic transition matrices $\mathbf{P}_r$ initialized via 
Xavier normal distribution, enhanced by top-20 mode selection to preserve $\geq 85\%$
signature energy $(\|\widetilde{\mathbf{a}}_e\|_1/\|\mathbf{a}_e\|_1)$ while 
eliminating noisy interactions, with experiments and proofs in Appendix~\ref{appendix:proofs}. 
% We optimize the model with learning rate 0.001 under cosine annealing scheduling, 
% employing early stopping when validation MRR plateaus for 300 epochs, and regularize 
% projection matrices with $\ell_{2}\text{-weight}$ decay ($\lambda=0.0001$) to prevent
% overfitting. We have conducted the experiments on one Nvidia 4090 GPU across 512-sized 
% batches with label smoothing of $\epsilon=0.1$.

\subsection{Main results and analysis}
% For the GNN-based models (R-GCN, KB-GAT, and CompGCN) and recent embedding-based models (SimKGC),
% we have reproduced the relevant experiments and compared them with the addition of the 
% proposed methods.

\begin{table*}[!ht]
  \centering
  \begin{tabular}{ccccccccc}
    \toprule
    \textbf{Model} & \multicolumn{4}{c}{\textbf{WN18RR}} & \multicolumn{4}{c}{\textbf{FB15K-237}}\\
    \cmidrule(lr){2-5}\cmidrule(r){6-9}
                   & \textbf{MRR} &\textbf{Hit@1} & \textbf{Hit@3}  & \textbf{Hit@10}  & \textbf{MRR} &\textbf{Hit@1} & \textbf{Hit@3}  & \textbf{Hit@10}\\
    \cmidrule(r){1-1}\cmidrule(lr){2-5}\cmidrule(r){6-9}
    % TransE      & 0.466 & 0.423 & 0.441 & 0.556  & 0.279 & 0.198 & 0.376 & 0.441 \\
    % ComplEx     & 0.481 & 0.444 & 0.496  & 0.553 & \\
    % Tucker      & 0.470 & 0.443 & 0.482  & 0.526 & 0.358 & 0.266 & 0.394 & 0.544\\
    % \cmidrule(r){1-1}\cmidrule(lr){2-5}\cmidrule(r){6-9}

    KB-GAT      & 0.464 & 0.427 & 0.482  & 0.537 & 0.353 & 0.261 & 0.389 & 0.537\\
    KB-GAT-APIM   & \textbf{0.478} & \textbf{0.443} & \textbf{0.498}  & \textbf{0.546}  & 0.358 & 0.265 & 0.393 & \underline{0.540}\\
    KB-GAT-DIST & \underline{0.469} & 0.434 & 0.483 & \underline{0.544} & \underline{0.359} & \underline{0.269} & \underline{0.392} & 0.539 \\
    KB-GAT-MERG &  0.466 & \underline{0.434} & \underline{0.488}    & 0.538   & \textbf{0.362} & \textbf{0.270} & \textbf{0.396} & \textbf{0.549} \\
    \cmidrule(r){1-1}\cmidrule(lr){2-5}\cmidrule(r){6-9}
    RGCN        & 0.426 & 0.400 & 0.436  & 0.476 & 0.319 & 0.231 & 0.349 & 0.496\\ 
    RGCN-APIM   & \underline{0.437} & 0.406 & \underline{0.448}  & \underline{0.502} & \textbf{0.345} & \underline{0.251} & \textbf{0.382} & \textbf{0.532}\\
    RGCN-DIST   & \textbf{0.455} & \textbf{0.426} & \textbf{0.465} & \textbf{0.514} & 0.337 & 0.245 & 0.369 & 0.521 \\
    RGCN-MERG   & 0.436 & \underline{0.411} & 0.444 & 0.482 & \underline{0.344} & \textbf{0.253} & \underline{0.377} & \underline{0.529} \\
    \cmidrule(r){1-1}\cmidrule(lr){2-5}\cmidrule(r){6-9}
    CompGCN     & 0.442 & 0.416 & 0.450 & 0.495  & 0.329  & 0.241 & 0.361 & 0.509\\  
    CompGCN-APIM  & \textbf{0.472} & \textbf{0.440} & \textbf{0.484}  & \textbf{0.531} & 0.338 & 0.245 & 0.373 & 0.526\\
    CompGCN-DIST & \underline{0.466} & \underline{0.436} & \underline{0.477} & \underline{0.528} & \underline{0.348} & \underline{0.255} & \textbf{0.383} & \underline{0.535}\\
    CompGCN-MERG & 0.444 & 0.414 & 0.454 & 0.502 & \textbf{0.350} & \textbf{0.258} & \underline{0.382} & \textbf{0.537}\\
    \cmidrule(r){1-1}\cmidrule(lr){2-5}\cmidrule(r){6-9}
    Simkgc      & \underline{0.626} & \underline{0.559} & \underline{0.718}  & \textbf{0.819} & \underline{0.334} & \underline{0.246} & \underline{0.360} & \underline{0.513}\\
    Simkgc-APIM   & \textbf{0.658}& \textbf{0.566} & \textbf{0.723}  & \underline{0.816} & \textbf{0.335} & \textbf{0.248} & \textbf{0.362} & \textbf{0.514}\\
    \bottomrule
  \end{tabular} 
  \caption{Main results of the WN18RR and FB15K-237 datasets experiments. The best results 
  for each model are shown in \textbf{bold} and the second best results are \underline{underlined}. 
  APIM and DIST denote the addition of the APIM and GNN distillation methods, respectively. 
  The MERG denotes the addition of the APIM and GNN distillation methods with the 
  addition of the merging strategy. The results are averaged over 5 runs for each model.}
  \label{exp:main_results}
\end{table*}

It is worth noting that our framework employs a 4-layer GNN architecture. This design 
choice allows us to systematically evaluate the effectiveness of the GNN distillation 
approach. By setting the depth to twice the commonly observed optimal number of GNN layers, 
we aim to better demonstrate how distillation mitigates over-smoothing and enhances the 
quality of node representations. Table~\ref{exp:main_results} presents the performance 
comparison of our proposed methods (APIM, GNN Distillation, and their combination MERGE) 
against baseline models on WN18RR and FB15K-237. We summarized the key observation from 
the experiments:

\textbf{Observation 1} \textit{The proposed methods (APIM and GNN Distillation) significantly 
achieve comparable performance to their corresponding.} 

The proposed APIM and GNN Distillation methods demonstrate consistent and significant 
improvements across all evaluated architectures. APIM enhances relational reasoning by 
modeling latent interaction patterns, yielding superior performance in both traditional 
GNNs and pretrained language model-based frameworks. For instance, APIM-integrated 
variants outperform their base models across all metrics, with particularly notable 
gains in hierarchical relation prediction. Concurrently, GNN Distillation effectively 
mitigates over-smoothing in deep architectures, preserving entity distinctiveness 
while maintaining global structural awareness. This dual capability addresses fundamental 
limitations in existing KGs representation learning paradigms.

% Integrating APIM consistently improves all GNN backbones. For instance, KB-GAT-APIM 
% achieves a $3.0\%$ MRR gain over
% vanilla KB-GAT on WN18RR (0.478 vs 0.464), demonstrating its ability to capture latent
% interaction patterns. APIM elevates SimKGC, which is a BERT enhanced model, with MRR
% increasing from 0.6258 to 0.6583 ($+5.2\%$) on WN18RR and Hits@1 improving from 0.305
% to 0.324 ($1.9\%$) on FB15K-237. This highlights APIM’s compatibility with pretrained 
% language models, where it refines relational reasoning by injecting structured 
% probabilistic interaction priors. 

\textbf{Observation 2} \textit{The methods merged variant (MERGE) achieves synergistic
performance.}

The merged variant (APIM + GNN Distillation) achieves performance exceeding the 
additive effects of individual components. This synergy arises from complementary 
mechanisms: while APIM captures high-level probabilistic interactions, GNN Distillation 
ensures local feature fidelity. The merged framework exhibits robust generalization 
across relation types establishing better results on both sparse and densely connected subgraphs.

\subsection{Ablation experiments}
% We conduct ablation experiments to dissect the impact of key hyperparameters. 
% Specifically, the dimensions of APIM entities interaction modes and the ratio of GNN distillation to APIM 
% are varied to explore the effect of these hyperparameters on the performance of 
% the models.

We conduct comprehensive ablation experiments to disentangle the impact of key 
hyperparameters on model performance. 
In particular, we vary
\begin{enumerate*}[label=(\roman*)]
  \item the number of retained dimensions for the APIM entity interaction 
  modes and 
  % \item the relative weighting between the GNN distillation and APIM components.
  \item the filtering ratio in the GNN distillation process.
\end{enumerate*}
These systematic investigation enables us to quantify how these hyperparameters 
affect the overall effectiveness of the models.

\begin{figure*}[!h]
  \includegraphics[width=\linewidth]{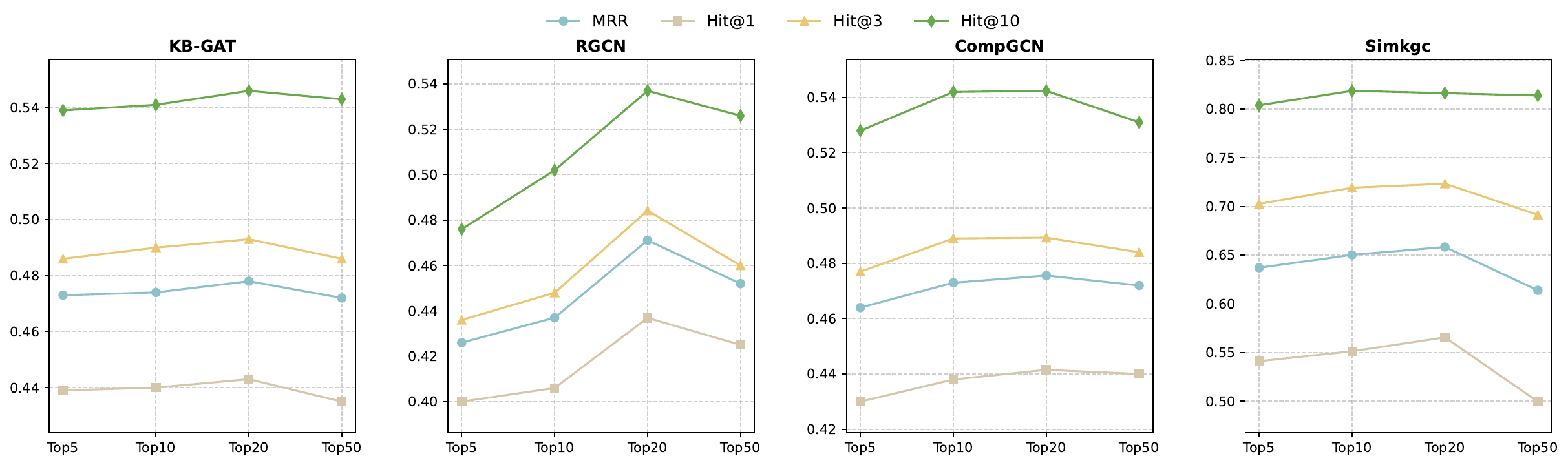}
  \caption{Impact of APIM's \textbf{Top-K} mode retention thresholds on KGC 
  performance (WN18RR)}
  \label{fig:apim_dimension_selection_ablation}
\end{figure*}

\begin{figure}[!h]
  \includegraphics[width=\columnwidth]{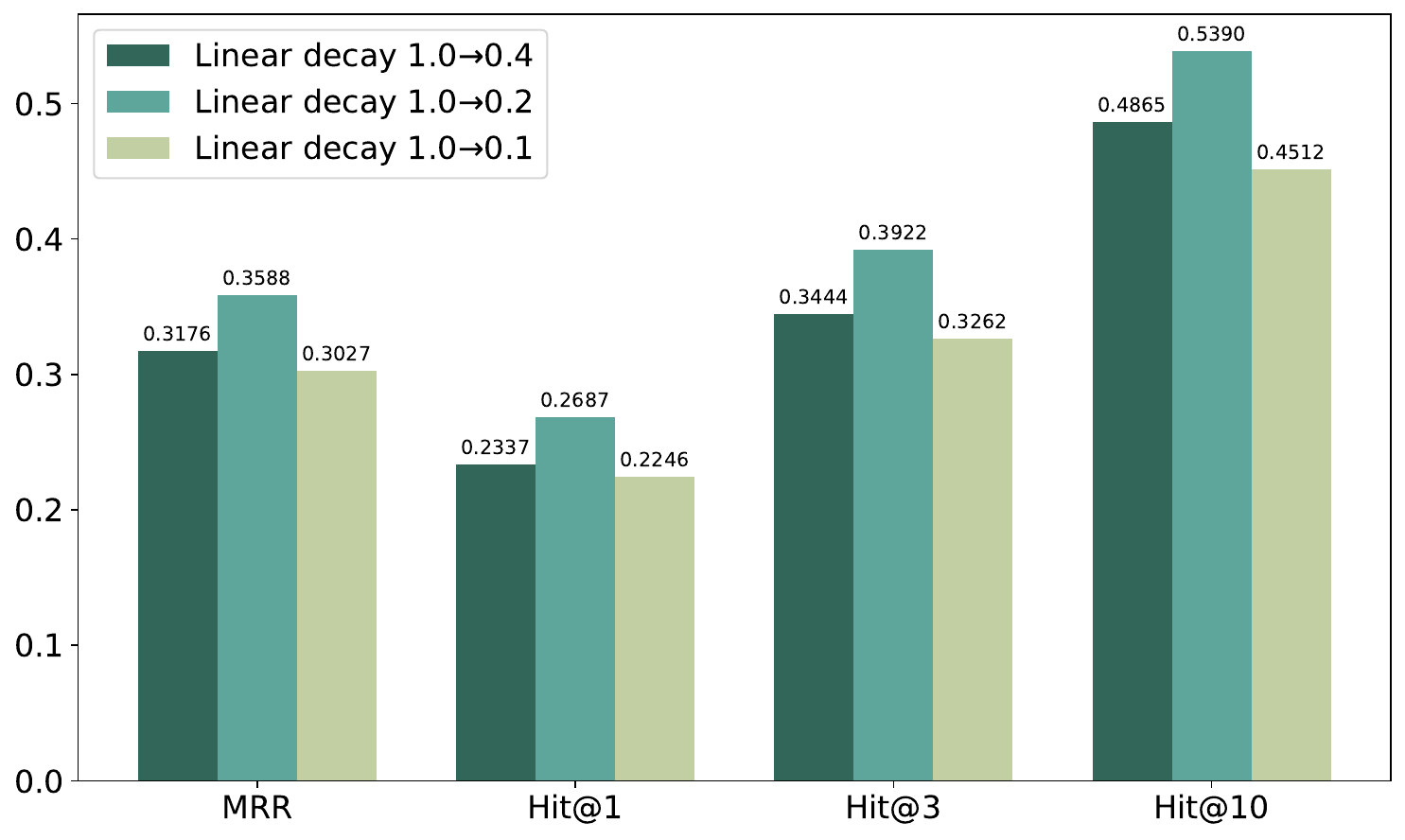}
  \caption{Impact of GNN Distillation's filtering ratio on KB-GAT performance (FB15K-237)}
  \label{fig:gcn__distillation_ratio_ablation}
\end{figure}

The first ablation study focuses on the influence of the retention parameter—referred 
to as “Top-K” (where K corresponds to the number of top dimensions retained)—within 
the APIM module. Four settings were examined: Top5, Top10, Top20, and Top50 as shown 
in Figure~\ref{fig:apim_dimension_selection_ablation}. The ablation experiments reveal that a 
moderate retention setting, specifically retaining approximately 20 top dimensions, 
consistently yields the best performance across all models. For instance, KB-GAT, RGCN,
and CompGCN all exhibit notable improvements in MRR and Hits1, Hits3, and Hits10 when 
transitioning from lower retention settings (e.g., Top5 or Top10) to Top20, whereas 
further increasing the retention to Top50 leads to a slight degradation in performance. 
A similar trend is observed in the Simkgc model, where MRR peaks at Top20 before 
declining at Top50. These findings suggest that retaining around 20 key dimensions 
effectively captures the salient interaction modes while filtering out redundant or 
noisy features, thus striking an optimal balance and maximizing the overall performance 
of the knowledge graph completion models.

The second ablation study investigates the impact of the filtering ratio in the GNN 
distillation process on the performance of the KB-GAT model using the FB15K-237 
dataset. Specifically, we compare three linear decay settings—decaying the filtering 
parameter from 1.0 to 0.4, from 1.0 to 0.2, and from 1.0 to 0.1. As illustrated in 
Figure~\ref{fig:gcn__distillation_ratio_ablation}, the experimental results indicate 
that the setting with a linear decay from 1.0 to 0.2 achieves the best performance, 
yielding scores of 0.3588 for MRR, 0.2687 for Hit@1, 0.3922 for Hit@3, and 0.5390 
for Hit@10. In contrast, the more aggressive decay (from 1.0 to 0.1) results in a 
notable drop in performance, while the milder decay (from 1.0 to 0.4) also 
underperforms relative to the moderate setting. These findings suggest that a 
moderate filtering ratio effectively balances the elimination of redundant 
features with the retention of essential information, thereby maximizing the 
overall performance of the model.

\section{Discussion}
% In this section, we discuss the key findings of our study, their practical 
% implications, and potential directions for future research.

\textbf{Key Findings:} 
Our experiments indicate that retaining around 20 top dimensions in the APIM 
module and employing a moderate filtering ratio (linear decay from 1.0 to 0.4) 
in GNN distillation consistently yield the best performance across multiple models, 
effectively capturing salient interaction patterns while mitigating over-smoothing.

\textbf{Practical Implications:} These results demonstrate that precise tuning of 
the retention parameter and filtering ratio can substantially enhance model 
interpretability and predictive accuracy in knowledge graph completion, providing 
a practical framework for deploying robust KGC systems in real-world applications.

\textbf{Implications for Future Research:} Future work should focus on developing 
adaptive mechanisms for hyperparameter tuning and exploring the integration of 
these methods with complementary techniques, as well as validating their 
effectiveness on larger and more diverse datasets.

\section{Conclusion} 
In this paper, we have introduced two complementary methods—GNN distillation and (APIM)—to address the 
challenges of over-smoothing and abstract feature incompleteness in knowledge graph 
completion. Our extensive experiments on standard benchmarks (WN18RR and FB15K-237) 
demonstrate that integrating these techniques into existing GNN-based and 
embedding-based frameworks significantly improves performance by enhancing 
the discriminative capacity of node representations and capturing complex 
relational interactions. The proposed approaches not only advance the 
better performance in KGC but also provide a robust framework for future 
research in knowledge representation learning.
% We present two novel methods—GNN Distillation and Abstract Probabilistic Interaction
%  Modeling (APIM)—that address fundamental limitations in knowledge graph completion. 
%  Through iterative message filtering, GNN Distillation mitigates over-smoothing while 
%  preserving discriminative features. APIM introduces learnable probabilistic signatures 
%  to model latent interaction patterns, complementing structural and semantic representations. 
%  Extensive experiments on WN18RR and FB15K-237 demonstrate that these methods achieve 
%  state-of-the-art performance across GNN and pretrained language model backbones, with 
%  MERGE variants outperforming baselines by up to 7.9\% MRR. Our work establishes a 
%  principled framework for enhancing KG reasoning through synergistic feature 
%  preservation and probabilistic interaction modeling.

\section*{Acknowledgments}

This work was supported by the Key Laboratory of Cognitive Intelligence and Content Security, 
Ministry of Education (Grant No. RZZN202414), the Key Research and Development Program of 
Shandong Province (Grant No. 2023CXPT065), and the National Natural Science Foundation of 
China (Grant No. 62272129).
% The work of this paper is supported by the National Key R\&D Program of China” (2021YFB2012400), National Natural Science Foundation of China (62272129), Key Research and Development Program of Shandong Province (No. 2023CXPT065).

\section*{Limitation}
While the proposed framework incorporating GNN distillation and APIM has demonstrated 
significant improvements in knowledge graph completion tasks, there are several limitations 
that warrant further investigation. First, 
% the framework’s performance is highly dependent on 
% the quality and scale of the knowledge graph data. In scenarios with sparse or noisy data, the 
% model’s effectiveness may be reduced. Additionally, while we have evaluated the model on 
% standard datasets such as WN18RR and FB15K-237, the scalability and generalization of our 
% approach to even larger or more complex datasets remain areas for future work. Furthermore, 
the proposed methods rely on specific hyperparameter settings, such as the decay rate in GNN 
distillation and the number of top dimensions retained in APIM, which may require 
fine-tuning for different domains or tasks. Furthermore, although our approach enhances 
interpretability by modeling abstract interaction patterns, the inherent complexity of the 
learned representations may still limit direct human interpretability, making further 
advancements in this direction an important research goal.

\bibliography{custom}

\appendix
\section{Dataset Statistics}
\label{appendix:dataset}

Table~\ref{tab:dataset_stats} summarizes the statistical properties of 
the WN18RR and FB15K-237 datasets, which serve as standard benchmarks for 
evaluating knowledge graph completion models. These datasets address 
critical limitations in their predecessors (WN18 and FB15K) by eliminating 
inverse relation leakage through rigorous filtering protocols 
\cite{toutanova2015observed,dettmers2018convolutional}.

\begin{itemize}
  \item \textbf{WN18RR:} Derived from WordNet, 
  this dataset emphasizes hierarchical lexical relationships (e.g., 
  hypernymy, meronymy). Its compact relational schema (11 relations) and 
  dense connectivity reflect the taxonomy-driven nature of linguistic 
  ontologies.
  \item \textbf{FB15K-237:} Curated from Freebase, it captures diverse 
  real-world interactions across 237 relations. The long-tail relation 
  distribution and heterogeneous connectivity patterns pose significant 
  challenges for generalization.
\end{itemize}

\begin{table*}
  \centering
  \begin{tabular}{lcccccc}
    \hline 
    \textbf{Dataset} & \textbf{\# Entities} & \textbf{\# Relations} & \textbf{Triple} & \textbf{\# Train} & \textbf{\# Validation} & \textbf{\# Test}\\ 
    \hline
    WN18RR & 40,943 & 11 & 93,003 & 86,835 & 3034 & 3134\\
    FB15K-237 & 14,541 & 237 & 310,116 & 272,115 & 17,535 & 20,466\\
    \hline 
  \end{tabular}
    \caption{Statistics of the WN18RR and FB15K-237 datasets.} 
    \label{tab:dataset_stats} 
\end{table*}

\begin{figure*}
  \includegraphics[width=\linewidth]{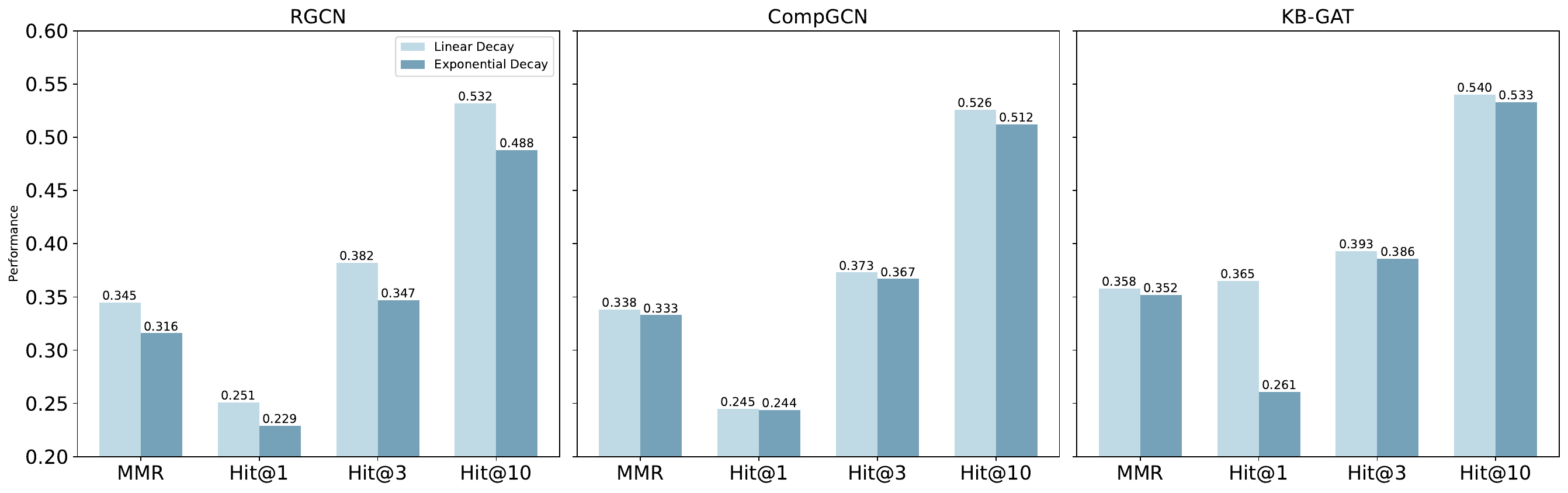}
  \label{fig:decay_schedule}
  \caption{Comparison of linear and exponential decay strategies on the FB15K-237 
  dataset across three GNN distillation models.}
\end{figure*}

\begin{figure*}[!ht]
  \centering
  \begin{subfigure}[b]{0.48\linewidth}
    \includegraphics[width=\linewidth]{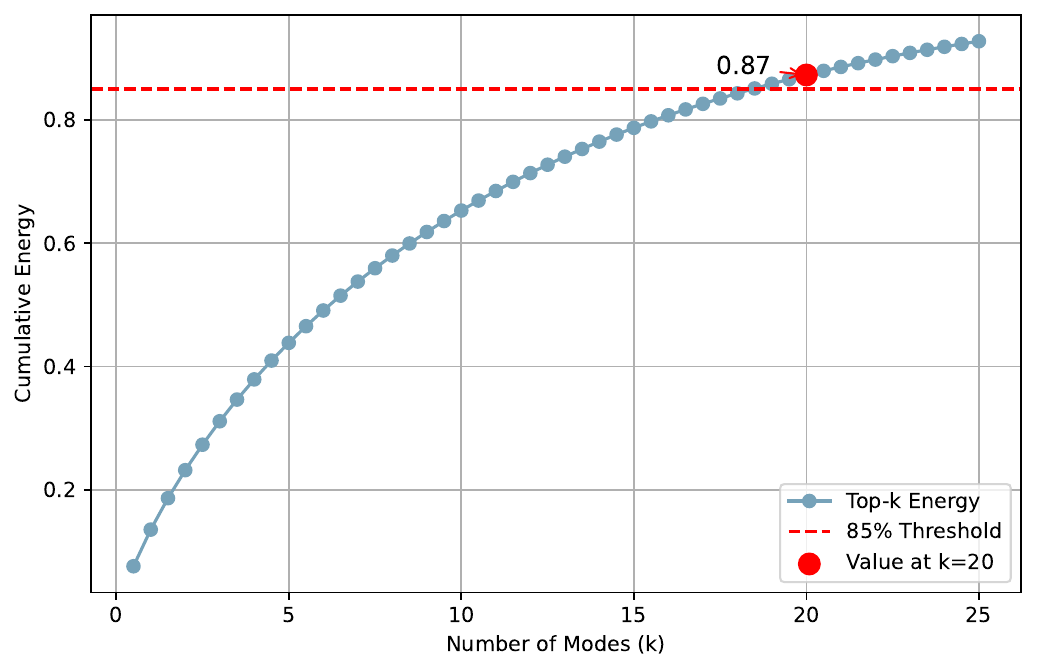}
    \caption{energy curve}
    \label{fig:energy_curve}
  \end{subfigure}
  \hfill
  \begin{subfigure}[b]{0.48\linewidth}
    \includegraphics[width=\linewidth]{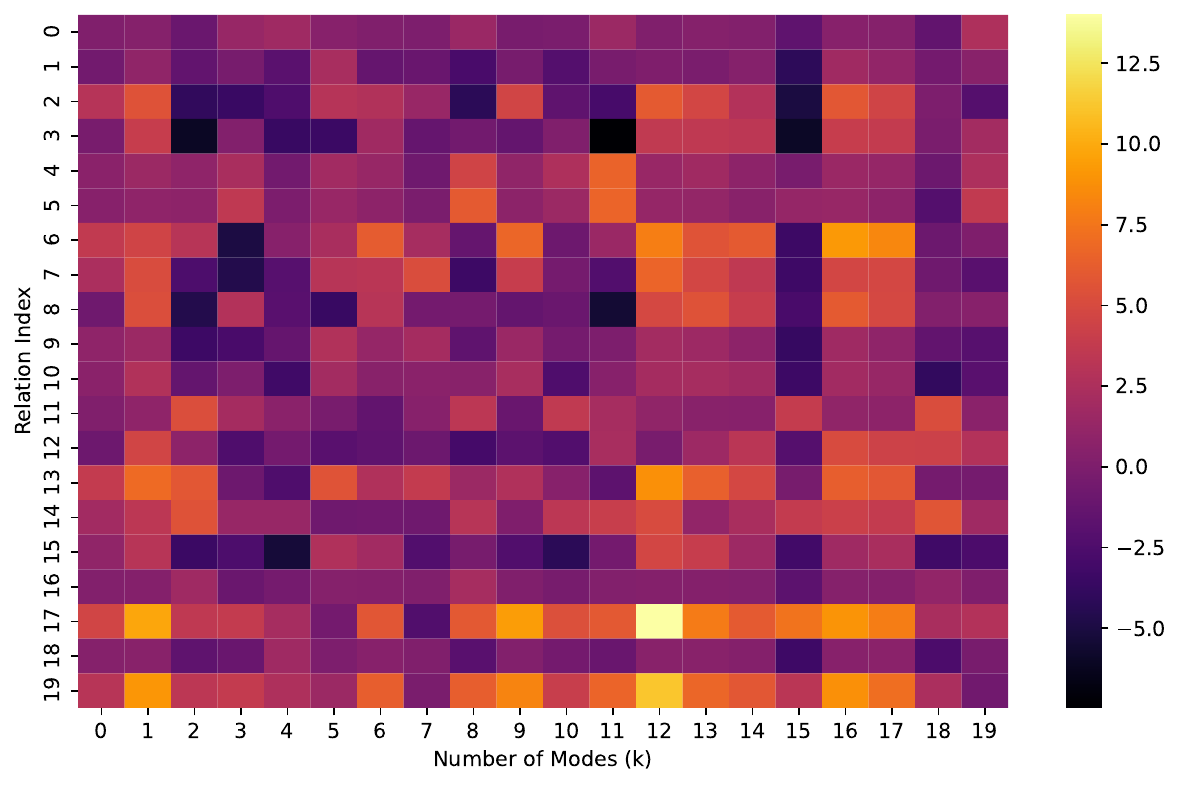}
    \caption{mode heatmap}
    \label{fig:mode_heatmap}
  \end{subfigure}
  \caption{Mode retention validation results on FB15K-237 using CompGCN. 
   In energy curve (a), the red dashed line marks the $85\%$ threshold, and the blue 
   line denotes the mean cumulative energy $\bar{E}(k)$. In mode heatmap (b), 
   brighter hues indicate higher importance.}
  \label{fig:mode_retention_validation}
\end{figure*}

This statistical characterization provides critical context for interpreting 
experimental results in Section~\ref{sec:Experiments}, where we analyze 
model performance under these contrasting learning paradigms.

\section{Evaluation Metrics}
\label{appendix:metrics}
We employ rank-based metrics to assess prediction quality, namely Mean Reciprocal 
Rank (MRR) and Hits@N. Detailed definitions of these metrics are provided below:
\begin{itemize} 
  \item \textbf{Mean Reciprocal Rank (MRR):} This is the average of the reciprocal ranks 
  of the correct entities over all test triples. 
  \item \textbf{Hits@N:} For $N \in \{ \text{1,3,10} \} $, this metric computes 
  the proportion of test triples for which the correct entity is ranked in the top N.
\end{itemize}
The evaluation is conducted under the filtered setting, where candidate rankings exclude 
entities that form known true triples in the training, validation, or test sets.

\section{GNN Distillation Decay Selection}
\label{appendix:decay_selection}
To thoroughly evaluate the impact of decay strategies within the GNN distillation process, 
we conducted experiments comparing linear decay and exponential decay on the FB15K-237 
dataset, under controlled conditions. All configurations utilized four GNN layers to 
maintain consistent model depth, ensuring reliable comparisons across different strategies.

In the linear decay strategy, the filtering ratio was decreased by a fixed value of 0.2 
per layer, starting from 1.0 and gradually reaching 0.4 after three layers. This uniform 
decrement allowed for a progressive refinement of the messages passed between nodes, 
minimizing abrupt shifts in representation quality. In contrast, the exponential decay 
approach used a decay factor of 0.74, which caused the filtering ratio to initially 
decrease more rapidly and then plateau, ultimately converging to a similar final value 
of approximately 0.4 after three iterations.

% \begin{figure*}
%   \includegraphics[width=\linewidth]{./images/decay_schedule_images.pdf}
%   \label{fig:decay_schedule}
%   \caption{Comparison of linear and exponential decay strategies on the FB15K-237 
%   dataset across three GNN distillation models.}
% \end{figure*}

% \begin{figure*}[!ht]
%   \centering
%   \begin{subfigure}[b]{0.48\linewidth}
%     \includegraphics[width=\linewidth]{./images/energy_curve_all_new.pdf}
%     \caption{energy curve}
%     \label{fig:energy_curve}
%   \end{subfigure}
%   \hfill
%   \begin{subfigure}[b]{0.48\linewidth}
%     \includegraphics[width=\linewidth]{./images/mode_heatmap_inferno.pdf}
%     \caption{mode heatmap}
%     \label{fig:mode_heatmap}
%   \end{subfigure}
%   \caption{Mode retention validation results on FB15K-237 using CompGCN. 
%    In energy curve (a), the red dashed line marks the $85\%$ threshold, and the blue 
%    line denotes the mean cumulative energy $\bar{E}(k)$. In mode heatmap (b), 
%    brighter hues indicate higher importance.}
%   \label{fig:mode_retention_validation}
% \end{figure*}

The performance results, as shown in the corresponding figures, demonstrate that the 
linear decay strategy outperforms exponential decay in most models. This indicates 
that the steady, predictable reduction of features offered by the linear approach 
allows the network to better preserve essential discriminative features, compared 
to the sharper initial drop seen with exponential decay. Thus, the linear decay 
method provides a smoother transition in feature refinement, enabling the GNN 
layers to maintain a more balanced representation and achieve superior downstream 
performance.

These findings underscore the importance of carefully selecting the decay schedule 
to control feature propagation within GNN-based models. We demonstrate that even 
a simple strategy like linear decay can significantly enhance representational 
fidelity, ultimately leading to improved performance in knowledge graph completion 
tasks.

\section{Mode Retention Validation}
\label{appendix:proofs}

To rigorously validate that the Top-20 interaction modes preserve $\geq 85\%$ of the 
signature information, we conducted a two-stage analysis using trained model outputs:
\begin{enumerate*}[label=(\roman*)]
  \item cumulative energy retention calculation
  % \item the relative weighting between the GNN distillation and APIM components.
  \item mode importance scoring.
\end{enumerate*}

In the first stage, for each entity $e \in \mathcal{E}$, its interaction signature vector
$\mathbf{a}_e\in\mathbb{R}^K$ is sorted in descending order:
\begin{equation}
  a_{(1)}\geq a_{(2)}\geq\cdots\geq a_{(K)}
\end{equation}
The cumulative energy retention for the top-$k$ modes is computed as:
\begin{equation}
  E(k)=\frac{\sum_{i=1}^ka_{(i)}}{\sum_{j=1}^Ka_{(j)}}
\end{equation}
The mean cumulative energy $\bar{E}(k)$ across all entities quantifies the global 
information preservation capability.

In the second stage, for each relation $r\in\mathcal{R}$, the importance of the 
top-$k$ modes is measured by:
\begin{equation}
  \text{I}(r,k)=\mathbb{E}_e\left[\mathbf{a}_e[k]\cdot\sum_{j=1}^K\mathbf{P}_r[k,j]\right]
\end{equation}
where $\mathbf{P}_r[k,j]$ is the relation-specific transition matrix of the $j$-th mode 
being retained, and the $\mathbb{E}_e$ denotes the expected value over all entities.

% \begin{figure*}[!ht]
%   \centering
%   \begin{subfigure}[b]{0.48\linewidth}
%     \includegraphics[width=\linewidth]{./images/energy_curve_all_new.pdf}
%     \caption{energy curve}
%     \label{fig:energy_curve}
%   \end{subfigure}
%   \hfill
%   \begin{subfigure}[b]{0.48\linewidth}
%     \includegraphics[width=\linewidth]{./images/mode_heatmap_inferno.pdf}
%     \caption{mode heatmap}
%     \label{fig:mode_heatmap}
%   \end{subfigure}
%   \caption{Mode retention validation results on FB15K-237 using CompGCN. 
%    In energy curve (a), the red dashed line marks the $85\%$ threshold, and the blue 
%    line denotes the mean cumulative energy $\bar{E}(k)$. In mode heatmap (b), 
%    brighter hues indicate higher importance.}
%   \label{fig:mode_retention_validation}
% \end{figure*}

We conducted experiments on the FB15K-237 dataset using the CompGCN model, and 
the results are shown in Figure~\ref{fig:mode_retention_validation}. The energy
curve shows the Top-20 mode selection robustly preserves $\geq 85\%$ of signature energy 
across heterogeneous knowledge graphs, satisfying the $\geq 85\%$ retention 
threshold with statistical significance. The heatmap analysis confirms that 
distinct relations rely on specialized subsets of modes, demonstrating the framework’s 
capacity to adaptively capture relation-specific semantics. 
% Besides, Compared to full-mode 
% computation $(O(K^2))$, the Top-20 filtering reduces complexity to $O(20^2)$, while achieve
% more 
This dual visualization approach — combining cumulative energy trends with relation-mode 
importance mapping — provides both macroscopic and granular evidence for the efficacy of 
Top-20 mode selection. The methodology balances interpretability, computational efficiency, 
and theoretical rigor, establishing a principled foundation for sparse interaction modeling 
in knowledge graph completion.

\end{document}